\documentclass[Times1COL,final]{WileyNJDv5}
\usepackage[utf8]{inputenc}
\usepackage{xcolor} 
\usepackage[most]{tcolorbox} 
\usepackage{times}
\usepackage{multicol}
\usepackage{amsmath}
\usepackage{array}
\usepackage{latexsym}
\usepackage{enumitem}
\usepackage{graphicx}
\usepackage{fontawesome5}
\usepackage{arydshln}
\usepackage{tikz}
\usepackage[T1]{fontenc}
\usepackage[utf8]{inputenc}
\usepackage{microtype}
\usepackage{inconsolata}
\usepackage{graphicx}
\usepackage[numbers]{natbib}
\articletype{Perspective}%

\startpage{1}

\usepackage{xcolor}
\makeatletter

\historydates{}
\doiheadtext{}

\renewcommand{\oddhead@titlepage@info}{}
\let\evenhead@titlepage@info\oddhead@titlepage@info

\renewcommand{\oddfoot@titlepage@info}{%
  \vbox{%
    \hsize\textwidth
    \noindent\hfill{\pagenumfont\thepage}\par
  }%
}
\let\evenfoot@titlepage@info\oddfoot@titlepage@info

\makeatother
\begin{document}

\definecolor{mytitlecolor}{HTML}{7769E0}
\title{On the use of foundation models in cognitive science}

\author[1]{Raj Sanjay Shah}
\author[2]{Alex Warstadt}
\author[3]{Michael Frank}
\author[1]{Sashank Varma}

\authormark{Shah \textsc{et al.}}
\titlemark{On the use of Foundation models in cognitive science}

\address[1]{\orgdiv{Interactive Computing}, \orgname{Georgia Institute of Technology}, \orgaddress{\state{Georgia}, \country{USA}}}

\address[2]{\orgdiv{Halıcıoğlu Data Science Institute and Dept. of Linguistics}, \orgname{University of California, San Diego}, \orgaddress{\state{California}, \country{USA}}}

\address[3]{\orgdiv{Dept. of Psychology}, \orgname{Stanford University}, \orgaddress{\state{California}, \country{USA}}}

\corres{Corresponding author: Raj Sanjay Shah \email{rajsanjayshah@gatech.edu}}

\abstract[Abstract]{
A host of recent studies have evaluated the cognitive and developmental alignment of Foundation Models (FMs). These investigations include evaluations of their correspondence to adult performance across a range of cognitive domains, as well as whether aspects of model training track children's cognitive development. However, using FMs as candidate cognitive models poses significant methodological and conceptual challenges. A key question underlies this effort: under what conditions does behavioral alignment justify treating FMs as explanatory models of cognition? In this paper, we articulate a four-stage inferential framework for evaluating FMs as cognitive and developmental models: adapting human experimental tasks to model-compatible formats, specifying linking hypotheses that map model outputs to human measures, evaluating behavioral correspondence, and comparing across candidate models or manipulations. We clarify the role of linking hypotheses in mapping model outputs to human behavioral measures, identify challenges that constrain alignment claims, and propose principles for theory-driven and comparative evaluation. Throughout, we argue that behavioral fit alone is insufficient. Alignment becomes scientifically meaningful only when embedded within explicit theoretical commitments, theory-diagnostic tasks, and systematic contrastive evaluation across candidate models.
}

\keywords{Cognitive modeling, Foundation models, Linking hypotheses, Cognitive science}

\jnlcitation{\cname{%
\author{Raj Sanjay Shah} and 
\author{Alex Warstadt} and 
\author{Michael Frank} and 
\author{Sashank Varma}
\ctitle{On the use of foundation models in cognitive science} \cjournal{\it } \cvol{}.}
}

\maketitle

\renewcommand\thefootnote{} 
\renewcommand\thefootnote{\fnsymbol{footnote}}
\setcounter{footnote}{1}

\section{Introduction}\label{sec1}

With the steadily improving performance of foundation models (FMs) \cite{anthropic2026claude, gemini31, gpt56}, researchers have increasingly explored their use as computational models of cognition \cite{piantadosi2023modern, mahowald2024dissociating, warstadt2022what, binz2025foundation}. Here, we use FMs as an umbrella term for pretrained, adaptable models that can support a range of tasks and modalities. Our scope includes large language models, multi-modal models, reasoning-oriented or post-trained models, as well as models trained under constrained regimes, such as BabyLM models. Across domains such as mathematical reasoning \cite{testolin2020numerosity, ahn2024large, liucogmath}, language comprehension \cite{duan2024hlb, cog_sci_garden_path, hu2024language}, conceptual understanding \cite{cog_sci_typicallity, bhatia2022transformer}, spatial reasoning \cite{ramakrishnan2024does, wang2024picture}, and analogical reasoning \cite{webb2023emergent, hu2023context, geiger2023relational}, FMs have been shown to reproduce behavioral signatures long documented by cognitive scientists. For example, recent work \cite{shah2023numeric} found that FMs exhibit the distance, size, and ratio effects characteristic of the human ``mental number line'' \cite{moyerTimeRequiredJudgements1967, parkman1971temporal, halberda2008individual}. Similarly, \citet{webb2023emergent} translated Raven's Progressive Matrices into symbolic digit matrices and reported FM performance comparable to humans. These findings have motivated the development of benchmarks aimed at systematically evaluating Model-human alignment \cite{coda2024cogbench, wang2025coglm}.

Researchers have also begun to use FMs to model cognitive development in children \cite{hosseini2022artificial, chang2022word, frank2023bridging, evanson2023language, ficarra2025distributional}. For example, \citet{Portelance2023PredictingAO} showed how language models can be used to predict the ages at which children acquire words. Rather than examining only model end states, \citet{shah2024development} evaluated intermediate training checkpoints to assess whether models' developmental trajectories in numerical ability, linguistic ability, conceptual understanding, and fluid reasoning parallel the developmental patterns observed in children. Similarly, \citet{tan2024devbench} explored developmental parallels by comparing the learning trajectories of vision-language models to both child and adult behavioral data. These and many other studies reflect a broadening shift in research goals towards examining their learning dynamics. Across both cognitive and developmental settings, we use `alignment' to mean systematic correspondences between model outputs and human behavioral measures \footnote{Our use of alignment is distinct from the one in AI safety, where it refers to aligning a system's goals, rewards, or behavior with human values or intentions.}.

\begin{figure*}[h]
\centering
\includegraphics[trim={0cm 0cm 0cm 0cm}, width=1\textwidth]{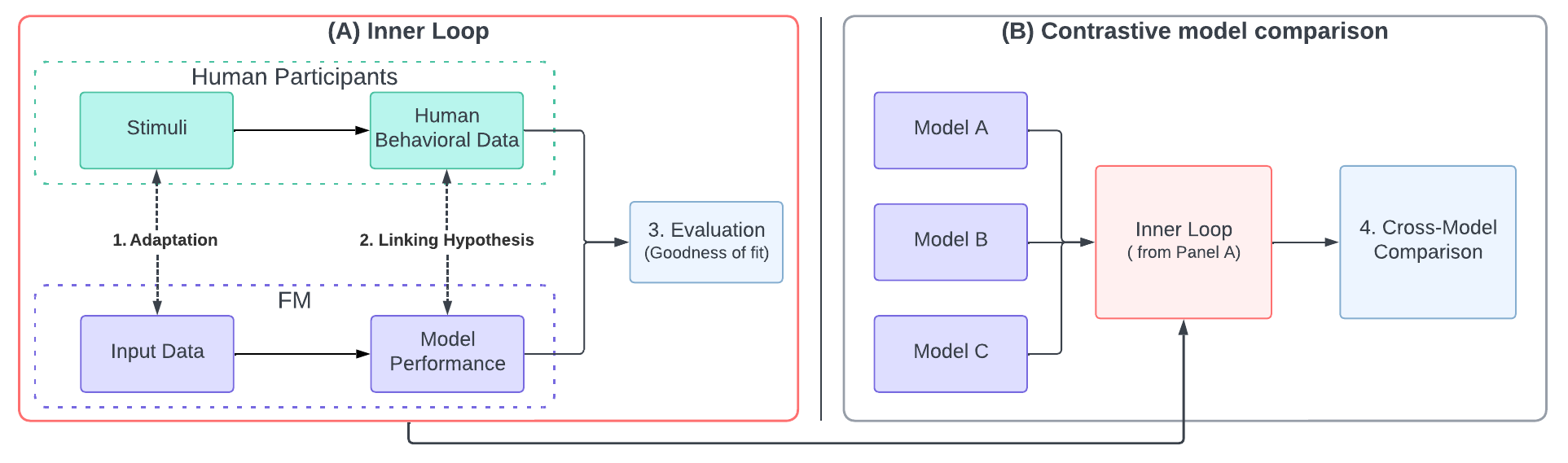}
\caption{A four-stage framework for evaluating FMs as cognitive models.
(A) The inner alignment loop establishes whether a given model reproduces human performance under explicit task adaptations and linking hypotheses (Stages 1-3).
(B) The outer contrastive loop evaluates multiple models or manipulations to identify computational features necessary for alignment (Stage 4).}
\label{fig:sufficiency}
\end{figure*}

In this paper, we ask: under what conditions does behavioral alignment justify treating foundation models as explanatory models of cognition? Because reproducing behavioral patterns does not by itself identify the mechanisms that generate them, researchers must distinguish explanatory cognitive models from behavioral proxies \cite{gershman2024have, van2024reclaiming}.
We propose an inferential framework for evaluating model-human alignment that makes three contributions. First, we present four evaluation stages (see Figure \ref{fig:sufficiency}), separating task adaptation, linking hypotheses, behavioral evaluation, and contrastive model comparison, and delineate how each stage constrains the interpretation of alignment. Second, we examine the interpretive role of linking hypotheses, the central theoretical commitment that connects model behavior to human performance measures. Third, we distill our proposals and analyses into guidelines for designing evaluations of the explanatory value (rather than the descriptive similarity) of FMs. Throughout, our goal is not to advocate uncritically for FMs as cognitive models but rather to clarify the conditions under which their alignment with human behavior becomes scientifically informative.

\section{A four-stage framework for evaluating FM alignment}\label{sec2}

Our proposed framework is directed toward research that uses computational models to test specific, theoretically motivated hypotheses \cite{frank2025cognitive}. Such efforts can support explanatory claims only when researchers make explicit what claim is being tested and what counts as ``alignment'' between model and human behavior. Without this structure, observed similarities risk being incidental. More strongly, explanatory relevance is strengthened when (1) the model's performance profiles match those of humans on the relevant dimensions specified by the hypothesis; (2) the model can predict human behavior in response to stimuli designed to test the hypothesis; and (3) the model's internal representations and processes can be interpreted as theoretical proposals about the underlying mechanisms. This perspective aligns with broader shifts toward model comparison and predictive adequacy in cognitive science \cite{yarkoni2017choosing}, while emphasizing that predictive success alone does not constitute explanation.

Our framework concerns the design and interpretation of model-human comparisons. It does not prescribe which architectures, objectives, or inductive biases constitute better cognitive models; these questions have been discussed elsewhere \cite{warstadt2022what}. Instead, it provides a structure for evaluating a theoretically motivated set of candidates. To operationalize these principles, we propose a four-stage framework for mapping between the performance of FMs and humans on cognitive tasks.

\begin{enumerate}
    \item \textbf{Adapt the experimental stimuli and task.} Human experimental materials must be translated into the input-output modalities of FMs so that the model and participants perform functionally equivalent tasks.

    \item \textbf{Specify the linking hypothesis.} Researchers must define how model outputs correspond to measurable aspects of human behavior. This mapping determines what counts as evidence of alignment, whether in terms of endpoint measures (e.g., accuracies, reaction times, error patterns) or developmental trajectories (e.g., age-of-acquisition curves, learning curves).
    
    \item \textbf{Evaluate correspondence with human performance:} Model performance must be compared to human data using appropriate statistical measures of goodness of fit. Importantly, alignment should be assessed not just in aggregate; emphasis must be placed on theoretically diagnostic variation, for example, across conditions, individuals, or stimuli.
    
    \item \textbf{Compare across candidate models and manipulations:} Demonstrating alignment in a single model provides, at most, proof of possibility. Explanatory value increases through contrastive evaluation: comparing architectures, training regimes, scales, or ablated variants to identify which computational components are \emph{necessary} to reproduce a behavioral signature. 
\end{enumerate}

As depicted in Figure \ref{fig:sufficiency}, Stages 1-3 form an inner alignment loop that establishes whether a model reproduces human performance under explicit task adaptations and linking hypotheses. Stage 4 functions as an outer, contrastive loop: by comparing architectures, training regimes, or ablated variants, researchers can identify which computational features are necessary for the correspondence. Related perspectives similarly emphasize that controlled manipulation and model comparison, rather than raw capability, are critical for making scientific progress with FMs \cite{ong2024gpt, warstadt2022what, rozner2026perturbationsimpleefficientadversarial}. Our framework systematizes such proposals.

\vspace{10pt}
\begin{tcolorbox}[
enhanced,
breakable,
title=Running Example: Digit-Matrix Analogical Reasoning,
colback=white,
colframe=mytitlecolor!60!black,
coltitle=white,
colbacktitle=mytitlecolor,
boxrule=0.6pt,
arc=4pt,
left=12pt,
right=12pt,
top=8pt,
bottom=10pt,
fonttitle=\bfseries,
attach boxed title to top left={yshift=-2mm, xshift=3mm},
boxed title style={
    arc=1.5pt,
    boxrule=0pt,
    colframe=mytitlecolor,
    colback=mytitlecolor,
}
]

\begin{minipage}{0.51\linewidth}
\citet{webb2023emergent, webb2025evidence} translated Raven's Progressive Matrices (RPM) problems into text-based ``digit matrices'' in which relational structure must be inferred to select the correct completion. This symbolic (vs. visual) representation (arguably) preserves the combinatorial and relational demands of fluid reasoning. 
They showed that FMs can consistently select (among multiple candidates) the correct completion (i.e., missing vector). The question is how such performance should be mapped to human reasoning processes.
\end{minipage}
\hfill
\begin{minipage}{0.47\linewidth}
    \centering
    \includegraphics[width=\linewidth]{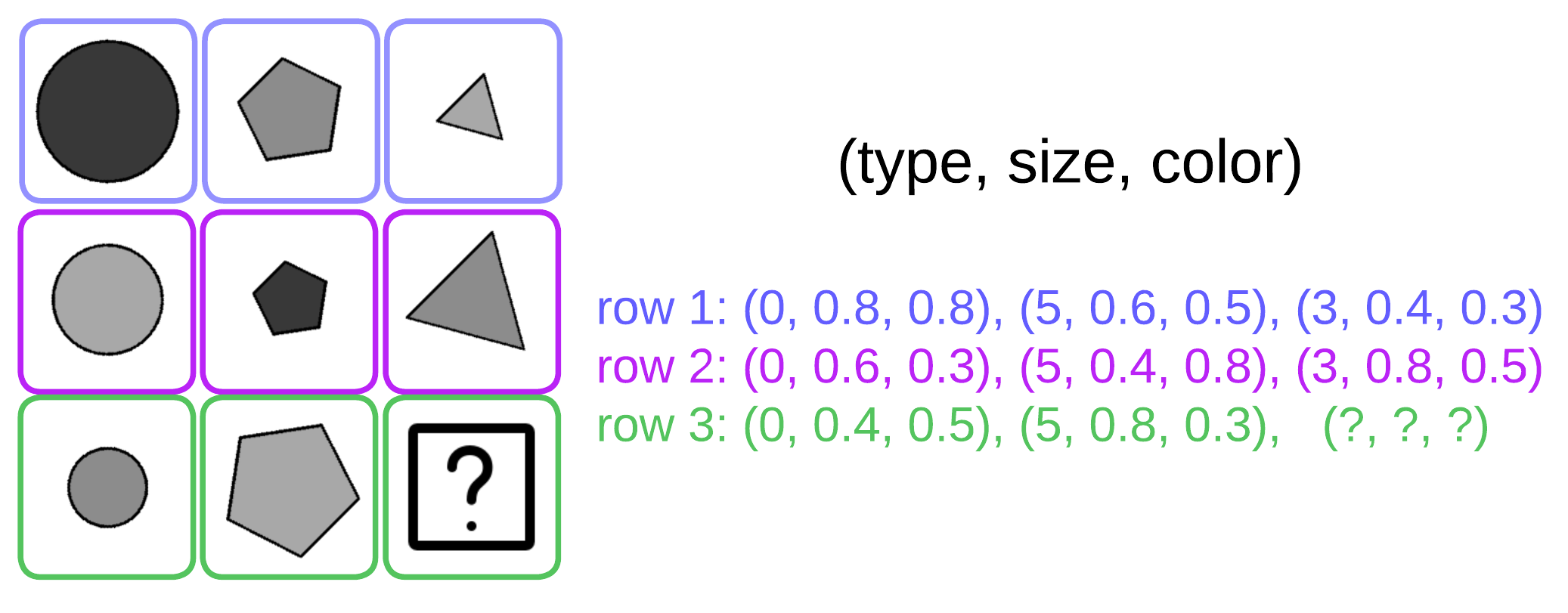}
    \captionof{figure}{ Adaptation of a visual RPM problem to a textual format. 
    }
    \label{fig:adaptation_new}
\end{minipage}

\vspace{10pt}

\textbf{Stage 1:}
To make this task compatible with the textual modality of FMs, \citet{webb2023emergent} converted each visual element into its symbolic attributes (e.g., type, size, color; see Figure \ref{fig:adaptation_new}), while preserving the relational structure of the original problem. This enables the model to reason over the same combinatorial relationships present in the original visual problem.

\textbf{Stage 2:}
Model behavior is then interpreted relative to human reasoning. One option is a \emph{prompting} strategy, in which the symbolic matrix and candidate answers are presented to the model exactly as they appear to human participants, and the model is asked to select the correct completion. Alternatively, a \emph{similarity-based} linking hypothesis could analyze the model's internal representations to test whether they encode the relational structure underlying the analogy.

\textbf{Stage 3:}
Model predictions are then evaluated against human behavioral data using goodness-of-fit measures. In this task, this involves comparing model accuracies, reaction times, and error patterns as a function of problem complexity to those observed in human participants.

\textbf{Stage 4:}

Finally, contrastive comparisons identify which computational properties support alignment. While \citet{webb2023emergent} primarily demonstrate that alignment is possible, subsequent studies (e.g., \citet{hu2023context}) vary models, problem representations, and prompting strategies to examine how these changes influence the model's ability to capture relational structure.
\end{tcolorbox}

\section{Linking Hypotheses: Mapping Model Performance to Human Performance}
\label{linking}

All three stages of the inner loop in Figure \ref{fig:sufficiency} are necessary for establishing a principled alignment.
Task adaptation (Stage 1) and performance evaluation (Stage 3) are often straightforward to operationalize.
By contrast, the choice of linking hypothesis fundamentally determines what counts as evidence of alignment and how model behavior is interpreted \cite{wurgaftcontext}.

The ``outputs'' of FMs are often quite different from the behavioral measures that cognitive scientists collect in their experiments. To bridge this gap, researchers use various \emph{linking hypotheses} to map model performance measures to human performance measures \cite{frank2025cognitive}. We focus here on several widely used linking hypotheses that illustrate how different mapping assumptions shape claims of alignment. These hypotheses are not neutral: each makes different theoretical commitments about what aspects of model computation map to cognition, thereby defining and potentially limiting how any observed alignment is interpreted. A strong apparent fit under one linking hypothesis might weaken or vanish under another. Consequently, alignment claims are not properties of models alone but of models evaluated under specific linking assumptions.

\subsection{Similarity}
One common linking hypothesis treats the geometry of a model's representational space as a proxy for the structure of human judgments or cognitive processes.  One form of this linking hypothesis concerns proximity: items that humans judge to be more similar or typical should occupy nearby regions of the model's representational space. Another concerns relational structure: stimuli or processes hypothesized to share a mechanism in humans might be represented through a common direction, rotation, or other transformation in the model's representational space \cite{hu2026representational}. Similarity-based linking has been used to model category typicality effects \cite{bhatia2022transformer, cog_sci_typicallity} and numerical comparison phenomena such as distance, size, and ratio effects \cite{shah2023numeric}. In these cases, distances in embedding space are mapped onto behavioral measures such as typicality ratings or response times.

Despite its apparent simplicity, this linking hypothesis requires substantial assumptions. It presupposes access to stable internal representations, which might not be available for closed-weight, commercial models. Computing a specific alignment score depends on a host of analytic choices, including layer selection, tokenization, and how category representations are defined (e.g., via label embeddings or averaged exemplars), each of which can materially alter similarity structure \cite{misra2021language, bhatia2022transformer}. Results further hinge on how representational geometry is characterized, with no \emph{a priori} guarantee that the chosen metric or transformation corresponds to psychologically meaningful structure \cite{richie2021similarity}. However, this greater flexibility also introduces additional analytic choices and makes it more difficult to determine which characterization of the representational space provides the appropriate linking hypothesis \cite{piantadosi2024concepts, lampinen2025representation}.

\subsection{Surprisal}

One way of quantifying the uncertainty of an FM's generation (given context) is in terms of its \emph{surprisal}, or the negative logarithm of a token's probability conditioned on the preceding tokens. A common linking hypothesis is that higher surprisal values correspond to longer human response times \cite{hale-2001-probabilistic, LEVY20081126}. FMs assign probabilities incrementally to text, enabling computation of surprisal for candidate continuations. In the digit-matrix analogy example \cite{webb2023emergent}, surprisal could be used to quantify the model's uncertainty over possible completions. If the correct completion has lower surprisal than alternatives, this can be interpreted as evidence that the model encodes the relevant relational structure. More generally, surprisal-based linking has successfully predicted graded effects in word learning, grammaticality judgments, and incremental sentence processing \cite{warstadt2020blimp, shain2024word}.

At the same time, adopting surprisal commits researchers to a particular relationship between predicted probability and processing difficulty. However, surprisal can be used for other purposes, such as testing hypotheses about how information is distributed across an utterance, without treating it directly as a behavioral measure. Moreover, surprisal is not model-independent: different models produce substantially different estimates, and larger models do not necessarily better predict human reading times \cite{oh2023transformer, kuribayashi2024psychometric}. It is also unclear where within a model surprisal should be measured, as estimates derived from intermediate layers can sometimes align better with human behavioral or neural responses than final-layer estimates \cite{kuribayashi2025large}. FMs are also highly sensitive to contextual framing, and modest variations in input structure can substantially alter surprisal values, complicating interpretation \cite{jiang2024llms, oh2023transformer}. These considerations suggest that the appropriateness of surprisal as a linking hypothesis is ultimately empirical and task-dependent.

\subsection{Prompting}

In prompting, an FM is instructed to follow the same instructions as human participants \cite{ivanova2025evaluate}. Thus, prompting is quite a ``proximal'' linking hypothesis, enabling direct comparison of the generations of FMs with the productions of humans, reducing (or even eliminating) the need for relatively ``distal'' (i.e., indirect) linking hypotheses \cite{patel2021mapping, webb2023emergent}. Because FM generation is probabilistic, repeated prompting can produce a distribution over responses, enabling direct comparison to distributions of human productions or choice probabilities. In the continuing digit-matrix analogy example \cite{webb2023emergent}, prompting would involve presenting the symbolic matrix and candidate completions exactly as shown to humans and asking the model to select the correct answer. Under this mapping, successful performance indicates that the model reproduces the observable behavioral outcome without requiring further analysis of intermediate representations (as with the similarity linking hypothesis) or interpretation of probabilities (as with surprisal). Prompting has been used successfully in studies of incremental sentence interpretation \cite{cog_sci_garden_path} and analogical reasoning \cite{lampinen2024language}.

Although prompting reduces the need for distal representational or probabilistic mappings, it shifts the interpretive risk to the elicitation procedure itself. Model performance can be highly sensitive to prompt wording and formatting \cite{guo2024understanding, binz2023using, ullman2023large}, raising questions about robustness. Responses might simply reflect instruction-following heuristics rather than the underlying cognitive mechanisms of interest, and systematic response biases (e.g., yes-response tendencies) can inflate apparent alignment \cite{hu2024language, dentella2023systematic}. Moreover, some research has questioned whether explicit metacognitive prompting (e.g., asking models to judge grammatical acceptability) faithfully captures the cognitive constructs of interest, especially for smaller-scale models \cite{hu2024auxiliary}.
Finally, models can fail to respect experimentally defined response options (e.g., producing explanations rather than forced-choice responses), complicating comparison to humans \cite{cai2024antagonistic}.

\subsection{Process-Trace Linking}

A major recent development in foundation models is the rise of reasoning-oriented systems that generate intermediate tokens before producing a final answer. For many open-weight systems, these traces and their token-level properties are directly accessible, making them an increasingly practical target for analysis. This has motivated a novel linking hypothesis: that properties of intermediate generation traces correspond to human processing dynamics.
For example, the length or structure of intermediate reasoning sequences (e.g., number of tokens, number of reasoning steps, or branching in generated explanations) might be mapped onto human response times or cognitive effort, with more extended traces reflecting increased difficulty.
In the digit-matrix analogy example \cite{webb2023emergent}, a model that generates longer intermediate reasoning sequences for more challenging Raven's problems could be interpreted as exhibiting graded processing analogous to human analogical reasoning (e.g., increasing response times). This proposal aligns with long-standing practices in cognitive psychology in which response times are treated as indirect measures of underlying cognitive processes \cite{luce1991response, anderson2009can, just2007organization}. More recently, research in large language models has proposed token-level measures of inference-time effort, such as reasoning-token budgets \cite{de2025cost, hu2026more}, deep-thinking ratios \cite{chen2026think}, and sequential computation budgets \cite{madaan2025rethinking}, as analogs of cognitive effort. 

Inferring cognitive processes from intermediate tokens raises interpretive challenges. As with response-time measures, differences in observable trace length do not necessarily imply differences in the targeted cognitive mechanism \cite{white2022need}. The causal impact of intermediate tokens on the final answer may be easy to overstate or misinterpret. The literature on chain-of-thought faithfulness shows that generated rationales can systematically misrepresent the factors that drive model responses: models might rationalize biased or already-determined answers, omit the true source of a decision, or expose answer-relevant information in hidden states before the reasoning trace is produced \cite{turpin2023language, cox2026decoding}. Intermediate tokens might reflect instruction-following conventions or stylistic artifacts (e.g., requesting outputs as JSON files) rather than intrinsic computational dynamics \cite{jaroslawicz2025many}. Thus, process-trace linking offers a promising yet still developing avenue for alignment \cite{vankov2026correlations, hu2026thinking}, and its validity hinges on whether intermediate token properties reliably capture meaningful cognitive computations.

\section{Challenges of using FMs as cognitive models}\label{sec4}

Using FMs as cognitive models raises several inferential challenges. The four-stage framework helps make these challenges explicit by identifying where alignment claims can break down and what kinds of evidence can strengthen them. These challenges do not undermine the potential scientific value of FMs; rather, they clarify the conditions under which model-human correspondences can support explanatory claims. In what follows, we articulate four challenges around a common structure: the inferential problem posed by FM-human alignment and a corresponding way forward. 

\subsection{Theoretical Underdetermination}

Cognitive science experiments are not neutral measurement exercises; rather, they are designed to adjudicate between competing hypotheses \cite{popper2014conjectures,varadarajan2025capturing}. Experimental conditions are constructed to be diagnostically informative, with certain conditions carrying outsize theoretical weight \cite{porada2024controlled}. This challenge arises most directly in \textbf{Stages 2 and 3} of the framework, where researchers specify linking hypotheses and evaluate behavioral correspondence. An FM might show high aggregate goodness of fit while failing on the critical contrasts that motivated the original experiment. Conversely, different linking hypotheses or analytic choices can yield different conclusions from the same model outputs. In both cases, behavioral correspondence alone does not uniquely determine which cognitive mechanism, if any, the model instantiates.

One way forward is to broaden the evidential base. Alignment claims are stronger when a model explains not only one dataset or effect, but a broader set of theoretically related phenomena (e.g., Centaur \cite{binz2025foundation}), including diagnostic contrasts across different stimuli, tasks, populations, and developmental stages. Broader coverage makes it harder for isolated empirical regularities, chance stimulus-level patterns, or flexible linking hypothesis choices to explain the observed fit, strengthening alignment.

\subsection{Mechanistic Opacity}

The second challenge is that FMs might reproduce human behavior without modeling the cognitive constructs that the experiment was designed to test. In Marr's terms, a model might align with humans at the computational level by solving the same task while diverging at the algorithmic level by relying on different representations, algorithms, or search strategies \cite{marr2010vision}. Mechanistic interpretability is a rapidly developing field attempting to identify functional subcomponents within models \cite{kar2022interpretability,ferrando2024primer,milliere2025interventionist}, and can therefore provide a secondary route through \textbf{Stages 2 and 3}: Researchers can ask whether model representations, activations, or processing dynamics correspond to theoretically relevant human measures. \textbf{Stage 4}-style comparisons, ablations, and causal interventions can then test whether those representational or computational features contribute to, and in some cases are necessary for, the observed behavioral alignment.

FMs make this cross-level inference difficult. Their internal components, such as layers, attention heads, and activation patterns, do not map cleanly onto cognitive constructs such as working memory, parsing strategy, conceptual similarity, or analogical mapping \cite{cichy2019deep, mcgrath2023can, rumelhart1986pdp}. This difficulty reflects the longstanding principle of multiple realizability, and as a result, behavioral alignment can overstate explanatory equivalence: two systems might produce similar outputs while relying on different internal computations \cite{putnam1967psychological,guest2023logical}. Although connecting lower-level model mechanisms to higher-level cognitive mechanisms remains difficult \cite{sharkey2025open}, such analyses are essential for evaluating whether the model actually implements the constructs proposed by the cognitive theory. Thus, mechanistic interpretability can contribute throughout the framework: it can support the formulation and evaluation of internal linking hypotheses in \textbf{Stages 2 and 3}, while \textbf{Stage 4} comparisons and interventions can test how the identified mechanisms contribute causally to behavioral alignment.

\subsection{Training and Developmental Mismatch}

The third challenge arises from differences between how FMs are trained and how humans learn. FMs are typically optimized via next-token prediction on large, static text corpora. Human learning is much richer by contrast. It unfolds interactively, under biological and environmental constraints, and is scaffolded by social interactions. Thus, differences in learning objectives and inductive biases might shape model representations in ways that diverge from human developmental trajectories \cite{warstadt2022what, sutton2019bitter, oh2025model}.

Developmental alignment presents an additional challenge. In principle, intermediate training checkpoints could be used to examine whether improvements in model performance track the developmental progressions observed in children \cite{frank2023bridging, shah2024development, evanson2023language}. Open training efforts that release these checkpoints \cite{biderman2023pythia, liu2023llm360, OLMo} create new opportunities for such analyses. However, differences in the quantity, modality, and ordering of training data complicate establishing such mappings \cite{cuskley2024limitations, charpentier-etal-2025-findings}. Even when model learning curves resemble developmental trajectories, such correspondences remain correlational and, by themselves, do not establish shared mechanisms of change.

Thus, claims of developmental alignment must be evaluated with particular caution. Stronger evidence for such claims requires causal manipulations of training regimes, data ordering, or architectural constraints to test whether such perturbations produce theoretically predicted changes in developmental trajectories. Recent efforts such as the BabyLM challenge \cite{choshen2026babylm} illustrate this opportunity by introducing developmentally motivated constraints on training data, exposure, and learning dynamics. These perturbations extend \textbf{Stage 4}-style contrastive evaluation beyond architectural ablations to test how training conditions shape developmental trajectories.

\subsection{Variability and Population-Level Alignment}

A final challenge concerns variability. FMs are typically trained on aggregate data and evaluated against average human performance. This emphasis mirrors the \emph{experimental} tradition in psychology, which often treats variability as noise. By contrast, the \emph{differential} tradition emphasizes individual differences as theoretically important \cite{carpenter1990one, Cronbach1957TheTD, underwood1975individual}. A growing body of work has examined whether FMs capture structured variation in human behavior \cite{mcduff2024cognitive, frank2025cognitive, fung2026individual}. Demonstrating alignment only at the group level means ignoring important dimensions of cognitive variability. At the same time, recent work has begun exploring the simulation of behavioral heterogeneity via persona-based prompting \cite{park2022social, tseng2024two, aher2023using}. Such research is still in its infancy, and conflicting findings show that further exploration is required \cite{milivcka2024large, salewski2024context}. Alignment claims that ignore individual differences in cognition risk overstating the generality of model-human correspondences.

More broadly, recent critiques caution that AI-based simulations might create illusions of generalizability, as the behavior of models can reflect the demographic and cultural biases of their training data rather than the diversity of human populations \cite{crockett2025ai}. A more informative approach is to treat variability as an observation to be explained rather than as noise to be averaged over. Alignment claims should ask whether models capture structured variation across individuals, groups, developmental stages, or contexts, and whether this variation follows theoretically predicted patterns. This reframes population-level heterogeneity as a target of explanation, not merely a source of error.

\section{Guidelines for Using FMs in Cognitive Science Research}

The four-stage framework presents a structured approach for evaluating FMs as candidate cognitive models. Here, we distill this framework into operational principles for research practice. These guidelines are not an exhaustive checklist of best practices, but a way of organizing inquiry so that alignment claims are developed with appropriate theoretical and empirical guardrails.

\paragraph*{1. Design evaluations around diagnostic measurement contrasts}

The goal of alignment is not simply to reproduce average performance, but to test whether models capture theoretically diagnostic distinctions. Experiments in cognitive science are typically designed to distinguish among competing hypotheses, with particular conditions carrying outsize explanatory weight \cite{popper2014conjectures}. When evaluating FMs, researchers should therefore prioritize contrasts that adjudicate among alternative accounts. As important as a model performing well overall (\textbf{Stage 3}) is whether its pattern of successes and failures discriminates among competing theories of the underlying cognitive mechanisms.

\paragraph*{2. Ensure that task adaptations preserve the diagnostic structure}

When translating human experiments into model-compatible formats, it is essential to preserve the logical distinctions of the original task. Adaptations should retain the combinatorial structure across conditions that makes an experiment theoretically informative (\textbf{Stage 1}). For example, in adapting a visual reasoning task into text, the adapted version should preserve the relational dependencies, distractor options, and graded difficulty that make the original task diagnostic. 
Superficially similar tasks might show low observed alignment because of inadvertently introduced formatting artifacts, interface constraints, or surface cues that change what the task measures.

\paragraph*{3. Triangulate across linking assumptions and behavioral measures}

Because alignment depends on how model outputs are mapped onto human data, when possible, researchers should evaluate performance under multiple relevant linking hypotheses (\textbf{Stage 2}). This corresponds to the experimental strategy of triangulating a phenomenon by investigating it using multiple dependent measures: response times, error rates, verbal self-reports, neuroimaging measures, etc. Different linking hypotheses reveal different aspects of model behavior. Convergent findings across linking strategies strengthen alignment claims, particularly when combined with \textbf{Stage 4}-style comparative evaluation across models and manipulations.

\paragraph*{4. Evaluate model alignment comparatively, not in isolation}

Model-human correspondence gains explanatory force when evaluated relative to alternative models and manipulated variants of the same model. Comparisons across architectures, training regimes, scales, or ablations help identify which computational features are necessary for reproducing behavioral signatures (\textbf{Stage 4}). In addition, comparisons with simpler statistical baselines (e.g., $n$-gram models \cite{michaelov2026n}, regression models \cite{boehm2018using}), and established cognitive models \cite{binz2025foundation} clarify what FMs uniquely contribute. Alignment is therefore inherently contrastive.

\paragraph*{5. Treat alignment as a starting point for causal or mechanistic investigation}

Close behavioral fit (\textbf{Stage 3}) is necessary but not sufficient for explanatory adequacy. Once correspondence has been established, the next step is to investigate whether the model's internal representations and computations support the same mechanisms posited by the target theory. This involves moving from whether the model aligns to how alignment is achieved within a model.
Such analysis might include ablation studies (\textbf{Stage 4}), representational probing, or examination of training dynamics to test whether disrupting specific components alters the observed behavior in theoretically predictable ways. Demonstrating alignment should thus be viewed as a starting point for more stringent investigation, rather than an endpoint.

\section{Conclusion}\label{sec5}
This paper has advocated for the careful use of Foundation Models as tools for investigating human cognition and its development. The scientific value of FMs lies not in outperforming humans or achieving leaderboard success, but in their potential to illuminate which computational principles are sufficient, and perhaps necessary, for reproducing characteristic patterns of human cognition. Here we distill and expand guidance from prior commentaries \cite{mcgrath2023can, frank2023bridging, warstadt2022what,mahowald2024dissociating} 
into a framework for establishing the alignment of FMs and humans on cognitive tasks (Figure \ref{fig:sufficiency}). Within this framework, alignment is not a property of models in isolation, but a claim embedded within explicit theoretical commitments, task adaptations, linking hypotheses, and comparative evaluation. We offer the four-stage framework not as a final, fixed prescription. As models evolve, training regimes diversify, and multimodal systems become more prominent, our framework will have to be adapted. This is important because FMs and their successors offer not just new tools for modeling behavior, but a new opportunity to sharpen our understanding of the structure, limits, and development of human thought.

\bmsection*{Acknowledgments}
We thank Cory Shain, Abhijit Mahabal, Ali Emami, Harsh Lalai, and Carrie Bruce for thoughtful feedback and helpful discussions on earlier versions of this manuscript.

\bibliography{custom}
\end{document}